\documentclass[10pt,twocolumn,letterpaper]{article}
\usepackage{cvpr}

\usepackage{graphicx}
\usepackage{amsmath}
\usepackage{amssymb}
\usepackage{booktabs}
\usepackage{multirow}
\usepackage{multicol}
\usepackage{bm}

\usepackage[pagebackref,breaklinks,colorlinks]{hyperref}

\usepackage[capitalize]{cleveref}
\crefname{section}{Sec.}{Secs.}
\Crefname{section}{Section}{Sections}
\Crefname{table}{Table}{Tables}
\crefname{table}{Tab.}{Tabs.}

\def\cvprPaperID{3152} 
\def\confName{CVPR}
\def\confYear{2023}

\begin{document}

\title{Learning Discriminative Representations for Skeleton Based Action Recognition}  

\author{Huanyu Zhou$^1$, Qingjie Liu$^{1,2,3,}$\thanks{Corresponding author}~, Yunhong Wang$^1$\\
$^1$State Key Laboratory of Virtual Reality Technology and Systems, Beihang University, Beijing, China\\
$^2$Zhongguancun Laboratory, $^3$Hangzhou Innovation Institute of Beihang University\\
{\tt\small \{zhysora, qingjie.liu, yhwang\}@buaa.edu.cn}
}

\maketitle

\begin{abstract}
Human action recognition aims at classifying the category of human action from a segment of a video. Recently, people have dived into designing GCN-based models to extract features from skeletons for performing this task, because skeleton representations are much more efficient and robust than other modalities such as RGB frames. However, when employing the skeleton data, some important clues like related items are also discarded. It results in some ambiguous actions that are hard to be distinguished and tend to be misclassified. To alleviate this problem, we propose an auxiliary feature refinement head (FR Head), which consists of spatial-temporal decoupling and contrastive feature refinement, to obtain discriminative representations of skeletons. Ambiguous samples are dynamically discovered and calibrated in the feature space. Furthermore, FR Head could be imposed on different stages of GCNs to build a multi-level refinement for stronger supervision. Extensive experiments are conducted on NTU RGB+D, NTU RGB+D 120, and NW-UCLA datasets. Our proposed models obtain competitive results from state-of-the-art methods and can help to discriminate those ambiguous samples. Codes are available at https://github.com/zhysora/FR-Head. 
\end{abstract}

\section{Introduction}


In human-to-human communication, action plays a particularly important role. The behaviors convey intrinsic information like emotions and potential intentions and thus help to understand the person. Empowering intelligent machines with the same ability to understand human behaviors is critical for natural human-computer interaction and many other practical applications, and has been attracting much attention recently.


\begin{figure}[t]
  \centering
	\includegraphics[width=\linewidth]{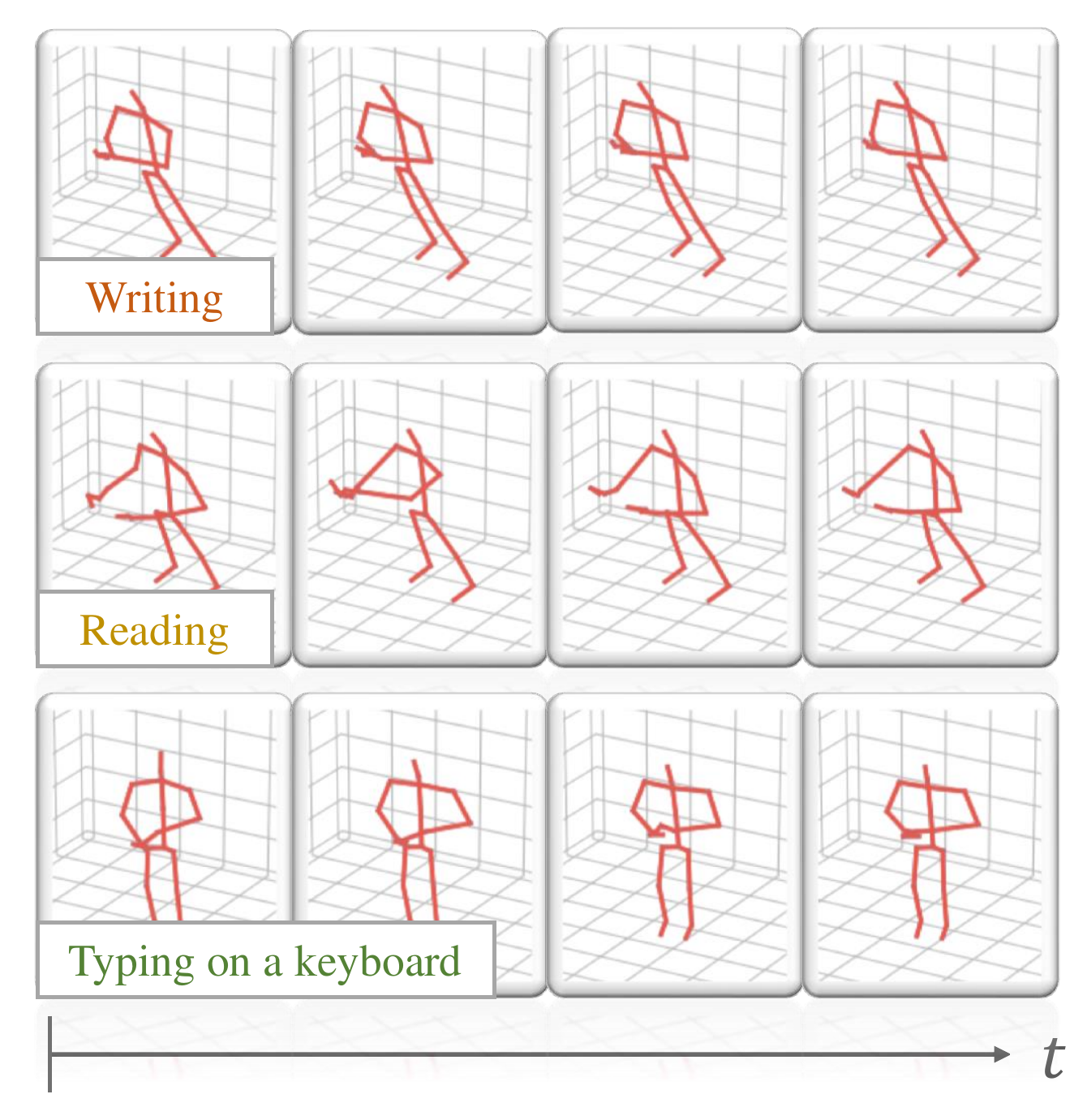}
   \caption{There are some actions that are hard to recognize because the skeleton representations lack important interactive objects and contexts, which make them easily confused with each other.}
   \label{fig:ambiguous sample}
\end{figure}


Nowadays, obtaining 2D/3D skeletons of humans has become much easier thanks to the advanced sensor technology and human pose estimation algorithms. Skeletons are compact and robust representations that are immune to viewpoint changes and cluttered backgrounds, making them attractive for action recognition. A typical way to use skeletons for action recognition is to build Graph Convolutional Networks (GCNs) \cite{ST-GCN-2018}. The joints and bones in the human body naturally form graphs, which make GCNs a perfect tool to extract topological features of skeletons. GCN-based methods have become more and more popular, with another merit that the models can be built lightweight and have high computational efficiency compared with models processing video frames.

However, using skeletons to recognize actions has some limitations. A major problem is that skeleton representation lacks important interactive objects and contextual information for distinguishing similar actions. As shown in Fig.~\ref{fig:ambiguous sample}, it is hard to distinguish ``Writing", ``Reading'' and ``Typing on a keyboard" based on the skeleton view alone. In contrast, a model can recognize them from RGB frames by focusing on the related items. These actions are easily confused with each other and should be given more attention. 

To alleviate this drawback, we propose a feature refinement module using contrastive learning to lift the discriminative ability of features between ambiguous actions. We first decouple hidden features into spatial and temporal components so that the network can better focus on discriminative parts among ambiguous actions along the topological and temporal dimensions. Then we identify the confident and ambiguous samples based on the model prediction during training. Confident samples are used to maintain a prototype for each class, which is achieved by a contrastive learning loss to constrain intra-class and inter-class distances. Meanwhile, ambiguous samples are calibrated by being closer to or far away from confident samples in the feature space. Furthermore, the aforementioned feature refinement module can be embedded into multiple types of GCNs to improve hierarchical feature learning. It will produce a multi-level contrastive loss, which is jointly trained with the classification loss to improve the performance of ambiguous actions. Our main contributions are summarized as follows:

\begin{itemize}
	\item We propose a discriminative feature refinement module to improve the performance of ambiguous actions in skeleton based action recognition. It uses contrastive learning to constrain the distance between confident samples and ambiguous samples. It also decouples the raw feature map into spatial and temporal components in a lightweight way for efficient feature enhancement.
	\item The feature refinement module is plug-and-play and compatible with most GCN-based models. It can be jointly trained with other losses but discarded in the inference stage.
	\item We conduct extensive experiments on NTU RGB+D, NTU RGB+D 120, and NW-UCLA datasets to compare our proposed methods with the state-of-the-art models. Experimental results demonstrate the significant improvement of our methods.
\end{itemize}

\section{Related Work}


\subsection{Human Pose Estimation}

Human pose estimation is an essential building block for a wide range of intelligent systems in fields such as AR, sports analysis, and healthcare, thus receiving much attention in recent years. Recent approaches leverage the temporal information of 2D pose sequences to alleviate the depth ambiguity in 3D poses \cite{2Dseqto3D1-2018, 2Dseqto3D2-2019, 2Dseqto3D3-2019, 2Dseqto3D4-2021, 2Dseqto3D5-2021}. Hossain \etal \cite{2Dseqto3D1-2018} tackle the task as a sequence-to-sequence problem and build RNNs to learn the mapping. Cai \etal \cite{2Dseqto3D2-2019} exploit the spatial-temporal relations from the 2D sequences via an encoder-decoder like GCN. Li \etal \cite{2Dseqto3D5-2021} use Transformer \cite{transformer-2017} to capture the long-range relationships in the 2D pose sequence.

\subsection{Skeleton Based Action Recognition}

Action recognition benefits a lot from human pose estimation. Early works treat the recognition as a sequence classification task. Su \etal \cite{SequenceModel2-2020} design an auto-encoder with RNNs to learn high-level features from the sequence. Another stream converts the skeleton sequence to image-like data using hand-crafted schemes \cite{ImageModel1-2018, ImageModel2-2019}. Duan \etal \cite{PoseC3D-2022} concatenate an RGB frame with a 2D skeleton heat map and use 3D CNNs to extract features. These works do not explicitly exploit the spatial structure of the human body.

The mainstream in this field is to use GCNs to extract the high-level features from skeletons since the joints and bones in the human body naturally construct a graph \cite{ST-GCN-2018, GEGCN-2019, MS-G3D-2020, CTR-GCN-2021}. In this way, the topology of the human body is fully exploited. Yan \etal \cite{ST-GCN-2018} is the first attempt to use GCNs for skeleton based human action recognition. They define the basic connections of spatial and temporal dimensions and introduce an efficient pipeline. Zhang \etal \cite{GEGCN-2019} build a two-stream architecture for both joint and bone modalities. Chen \etal \cite{CTR-GCN-2021} improve the design of GCNs in \cite{ST-GCN-2018} and propose to dynamically learn different topologies and effectively aggregate joint features in each channel.

\begin{figure*}[t]
  \centering
   \includegraphics[width=\linewidth]{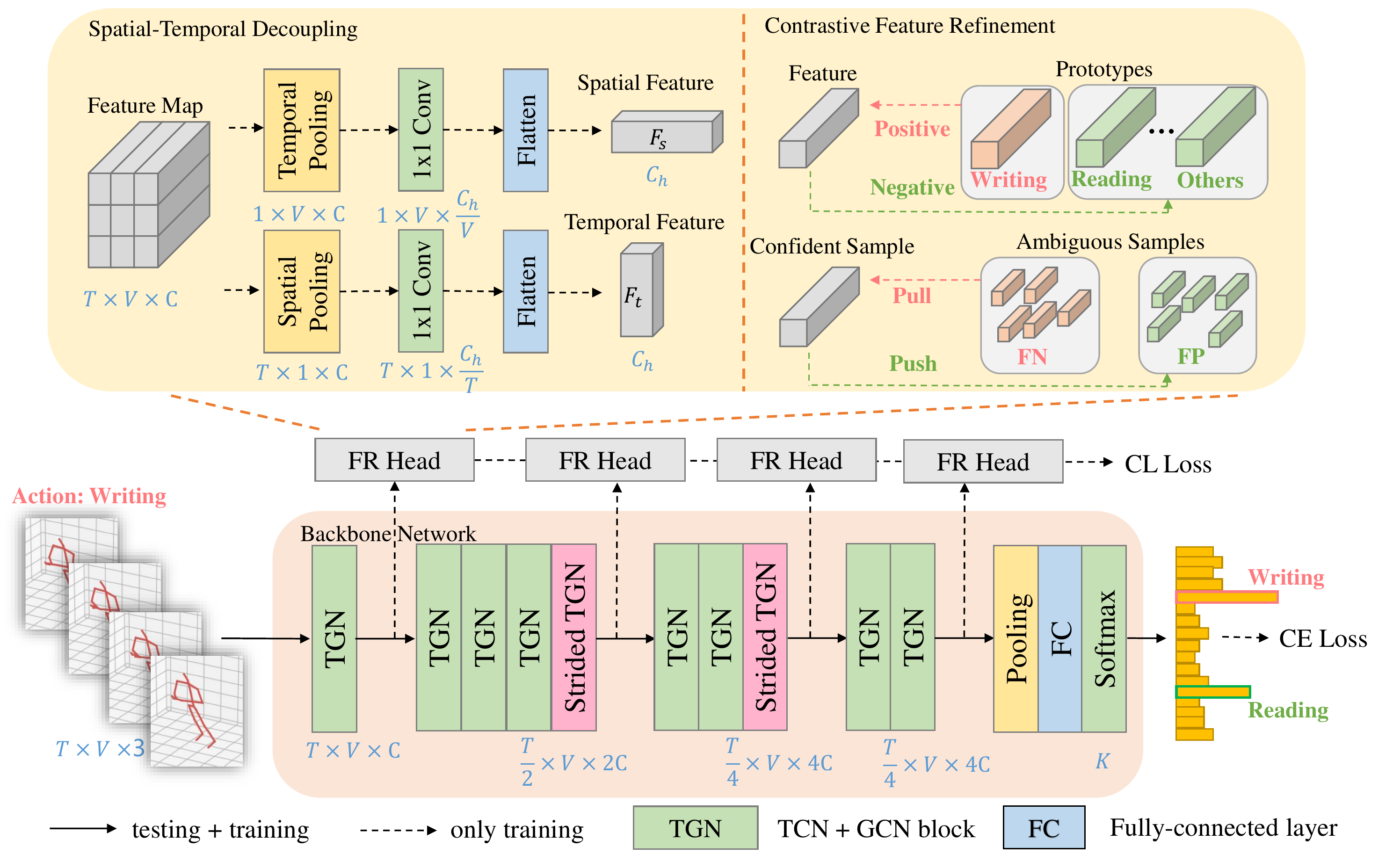}
   \caption{overview of the proposed method.}
   \label{fig:overview}
\end{figure*}

\subsection{Contrastive Learning}

Recently, contrastive learning has achieved remarkable progress in diverse fields. Typically, contrastive learning requires generating a set of transformed versions (or ``views") of an image using data augmentations, then training the network to distinguish the different views of the image. Chen \etal \cite{SimCLR-2020} explore the strategies of data augmentations and use a huge batch size to obtain the enhanced representation. He \etal \cite{MoCo1-2020, Moco2-2020, Moco3-2021} realize contrastive learning in a more efficient way using a momentum encoder and a dynamic queue. Wang \etal \cite{Renovate-2022} define the positive and negative samples in a supervised manner. They design a metric function loss to calibrate these misclassified feature representations for better intra-class consistency and segmentation performance. Inspired by this, our work tries to use a similar spirit to refine the skeleton representations for ambiguous actions.

\subsection{Ambiguous Sample}

Most of the recognition tasks for solving ambiguous samples focus on fine-grained image classification. For example, Lin \etal \cite{Lie-Group2015} perform bilinear pooling on the representations of two local patches to learn the discriminative feature. Dubey \etal \cite{dubey2018pairwise} model similarity between image pairs and leverage metric learning to improve the feature distributions. Zhuang \etal \cite{zhuang2020learning} design a module to adaptively discover contrastive cues from a pair of images and attentively distinguish them via pairwise interaction. However, we have not found works that aim at solving skeleton-based action recognition.

\section{Methodology}

We now give the details of our method. The overview of it is depicted in Fig.~\ref{fig:overview}.

\subsection{Backbone}

The input of our model is a sequence of skeletons with a shape of $T \times V \times 3$, which means $T$ frames of $V$ joints in a 3D space. We build our approach on \cite{CTR-GCN-2021}. However, we will show in the experimental section that it can improve any GCNs. The backbone consists of 10 basic units, termed TGN. TGN is constructed by a series of Temporal CNNs (TCNs) and Graph Convolution Networks (GCNs). Concretely, TCNs extract the temporal features by imposing 1D CNNs on the temporal dimension; GCNs extract the spatial features with a learnable topological graph defined on the spatial dimension. Note that two of the basic units are strided TGNs implemented by strided 1D CNNs. They are used to generate multi-scale features by decreasing the temporal dimension while increasing the channel dimension. Then, a pooling layer is applied to get the 1D high-level feature vectors. Finally, a fully-connected (FC) layer with softmax activation maps the feature to a probability distribution of $K$ candidate categories.

It is noted that the detailed implementation of the backbone is not the main concern of our method. The implementation of the basic unit can be replaced by any other GCN-based networks like \cite{ST-GCN-2018, 2s-AGCN2019}. 

\subsection{Feature Refinement Head}

Our main idea is to improve the performance of the skeleton based model on ambiguous actions that are quite similar and easily misclassified. To achieve this, we propose a plug-and-play module to optimize multi-level features within the backbone network, termed Feature Refinement Head (FR Head). It first decouples the hidden feature maps into spatial and temporal components and then applies a contrastive learning loss with global class prototypes and ambiguous samples. It is worth noting that the proposed FR Head is added only for training. There is no additional computational cost or memory consumption during inference.

\subsubsection{Multi-Level Feature Selection}

To learn more discriminative feature representations, we divide the backbone into four stages, respectively at the 1st, 5th, 8th, and last layer of TGN, and impose a FR Head on each of them. The 5th and the 8th layers employ a strided operation. Each FR Head refines the corresponding hidden features by calculating a contrastive learning (CL) loss, whose details will be discussed in Section.~\ref{sec:: CL learning}. To balance the different levels, we add a weighting parameter for each stage and the multi-level CL loss can be defined as a weighted average sum:

\begin{equation}
\label{eq:L_CL}
	\mathcal{L}_{CL} = \sum_{i=1}^{4} \lambda_i \cdot \mathcal{L}_{CL}^i 
\end{equation} 
where $\mathcal{L}_{CL}$ is the multi-level CL loss, $\lambda_i$ is the hyper-parameter to control stage $i$ and $\mathcal{L}_{CL}^i$ is the local CL loss calculated by stage $i$. 


\subsubsection{Spatial-Temporal Decoupling}

Due to the complexity of human activities, coarse modelling features will lead to confusion between ambiguous actions with similar spatial appearances or temporal transformations. 

For example, ``put sth. into a bag'' can be easily distinguished from ``take sth. out of a bag'' using temporal clues. However, compared to the ``reach into pockets'', more concentrations on the spatial information are required. Therefore, we propose a spatial-temporal decouple module that mines the spatial and temporal information simultaneously to improve the discriminative ability of action representations.


As Fig.~\ref{fig:overview} describes, the raw feature map is fed into two parallel branches for efficient feature enhancement. Concretely, each branch comprises a spatial/temporal pooling layer which only keeps the average value of the related dimension and a $1 \times 1$ convolution layer which is used to squeeze the feature to a fixed size. Then the output feature is flattened to a unified representation with the channel size of $C_h$. Finally, a CL loss is added on top of each branch. We accomplish the Spatial-Temporal Decoupling feature refinement by summing losses from the two branches:

\begin{equation}
	\mathcal{L}_{CL}^i = \text{CL}(F_s^i) + \text{CL}(F_t^i)
\end{equation} 
where $F_s^i$ and $F_t^i$ stand for the spatial feature and the temporal feature of stage $i$, respectively.  $\text{CL}(\cdot)$ is the function to calculate the CL loss with a specific feature vector.

\begin{figure}[t]
  \centering
   \includegraphics[width=\linewidth]{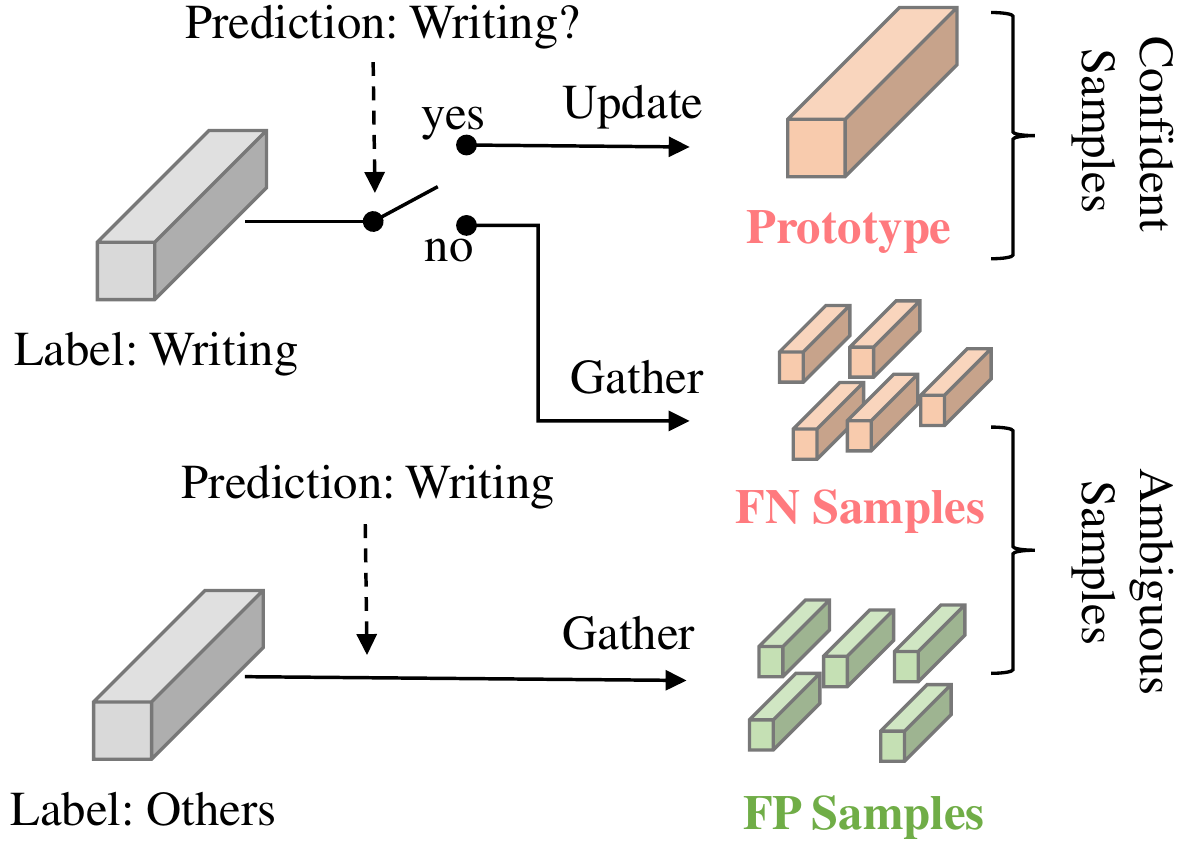}
   \caption{Discovering the confident and ambiguous samples for action ``writing''.}
   \label{fig:Gathering}
\end{figure}

\subsubsection{Contrastive Feature Refinement}
\label{sec:: CL learning}

As Fig.~\ref{fig:overview} shows, we conduct the feature refinement in a manner of contrastive learning. The idea is inspired by \cite{Renovate-2022}. Each sample will be refined by both its ground truth actions and other ambiguous actions. 

\textbf{Confident Sample Clustering}.
Given an action label $k$, if a sample is predicted correctly, namely as a True Positive (TP), we consider it a confident sample to distinguish it from ambiguous samples. Apparently, features from confident samples tend to have better intra-class consistency. As Fig.~\ref{fig:Gathering} shows, we gather those features to update the global representation (\ie, prototype) of the corresponding classes via exponential moving average (EMA). Assuming $s_{TP}^k$ is the set of confident samples for action $k$ in a batch and its size is $n_{TP}^k$, the EMA operation can be defined as:

\begin{equation}
	P_k = (1 - \alpha) \cdot \frac{1}{n_{TP}^k} \sum_{i \in s_{TP}^k} F_i + \alpha \cdot P_k 
\end{equation} 
where $P_k$ is the prototype of action $k$, $F_i$ is the feature extracted from sample $i$. $\alpha$ is the momentum term, we set it to $0.9$ by experience.
Over the training procedure, the prototype becomes a stable estimation of the clustering center for action $k$. It is capable of refining the feature of a newly arrived sample. Each sample should be close to the related prototype while far away from other prototypes. The distance between two feature vectors is defined as $\text{dis}(\cdot, \cdot)$, which is implemented by cosine distance:
\begin{equation}
	\text{dis}(\bm{u}, \bm{v}) = \frac{\bm{u} \bm{v}^T}{||\bm{u}||_2 ||\bm{v}||_2}	
\end{equation}
where $\bm{u}, \bm{v}$ stand for 1D vectors. $||\cdot||_2$ is $L_2$ norm.


\textbf{Ambiguous Sample Discovering}.
To discover the ambiguous samples during the training stages, we gather the misclassified samples which tend to be quite similar to other categories. Given an action label $k$, there are two types of ambiguous samples. If a sample of action $k$ is misclassified as other categories, it is termed as False Negative (FN). If a sample of other categories is misclassified as action $k$, it is termed as False Positive (FP). Supposing $s_{FN}^k, s_{FP}^k$ are sets of FN and FP samples for action $k$ and the sizes of them are $n_{FN}^k, n_{FP}^k$. As fig.~\ref{fig:Gathering} shows, We gather those samples in a batch and calculate the mean values as the center representations:
\begin{equation}
	\bm{\mu}^k_{FN} = \frac{1}{n^k_{FN}} \sum_{j \in s_{FN}^k} F_j,~~\bm{\mu}^k_{FP} = \frac{1}{n^k_{FP}} \sum_{j \in s_{FP}^k} F_j
\end{equation}
where $\bm{\mu}^k_{FN}, \bm{\mu}^k_{FP}$ stand for the center representation of FN and FP samples of class $k$. Note that, we do not maintain a global representation for those ambiguous samples because the prediction of those samples is not stable in the training stage and the amount is much less than TP samples. 

\textbf{Ambiguous Sample Calibration}.
To calibrate the prediction of ambiguous samples, we take confident sample $i$ of action $k$ as the anchor and calculate an auxiliary term in the feature space. For those FN samples which should be classified as action $k$, a compensation term $\phi_i$ is introduced:
\begin{equation}
	\phi_i = \left\{
	\begin{aligned}
		1 - \text{dis}(F_i, \bm{\mu}^k_{FN}) &, \mbox{if } i \in s^k_{TP} \mbox{ and } n^k_{FN} > 0;\\
		 0 &, \mbox{otherwise}.\\
	\end{aligned}
	\right.
\end{equation}
By minimizing the compensation term $\phi_i$, FN samples are supposed to be closer to the confident sample in the feature space. When there are no FN samples or the cosine distance converges to 1, $\phi_i$ reaches the minimum value $0$. This may motivate the model to correct these ambiguous samples as action $k$.

On the other hand, for those FP samples which belong to other categories, a penalty term $\psi_i$ is introduced:
\begin{equation}
	\psi_i = \left\{
	\begin{aligned}
		1 + \text{dis}(F_i, \bm{\mu}^k_{FP}) &, \mbox{if } i \in s^k_{TP} \mbox{ and } n^k_{FP} > 0;\\
		 0 &, \mbox{otherwise}.\\
	\end{aligned}
	\right.
\end{equation}
Similarly, the penalty term $\psi_i$ penalizes the distance between the FP samples and the confident samples in the feature space. When there are no FP samples or the cosine distance converges to -1, $\psi_i$ reaches the minimum value $0$. This may prevent the model from recognizing these ambiguous samples as action $k$. 

Finally, taking sample $i$ as an anchor, the proposed CL loss function can be defined as:
\begin{equation}
\begin{split}
	\text{CL}(F_i) = - \text{log} \frac{ e^{ \text{dis}(F_i, P_k) / \tau - (1 - p_{ik}) \psi_i }  }{ e^{ \text{dis}(F_i, P_k) / \tau - (1 - p_{ik}) \psi_i } + \sum_{l \neq k} e^{ \text{dis}(F_i, P_l) / \tau}  }  \\
	- \text{log} \frac{ e^{ \text{dis}(F_i, P_k) / \tau - (1 - p_{ik}) \phi_i }  }{ e^{ \text{dis}(F_i, P_k) / \tau - (1 - p_{ik}) \psi_i } + \sum_{l \neq k} e^{ \text{dis}(F_i, P_l) / \tau}  }  
\end{split}
\end{equation} 
where $p_{ik}$ is the predicted probability score of sample $i$ for class $k$. It means that the TP samples with weaker confidence get stronger supervision from those ambiguous samples.

\subsection{Training Objective}

We use Cross-Entropy (CE) loss to train our network:
\begin{equation}
\label{eq:L_CE}
\mathcal{L}_{CE} = - \frac{1}{N} \sum_i 	\sum_c y_{ic} \mbox{log}(p_{ic})
\end{equation}
where $N$ is the number of samples in a batch. $y_{ic}$ is the one-hot presentation of the label of sample $i$. If and only if $c$ is the target class of sample $i$, $y_{ic} = 1$. $p_{ic}$ is the probability score of sample $i$ belonging to class $k$ predicted by the network.

Finally, CE loss is combined with our proposed multi-level CL loss to form the full learning objective function:
\begin{equation}
	\mathcal{L} = \mathcal{L}_{CE} + w_{cl} \cdot \mathcal{L}_{CL}
\end{equation}
where $\mathcal{L}_{CL}$ and $\mathcal{L}_{CE}$ are defined in Eqs.~\ref{eq:L_CL} and \ref{eq:L_CE}. $w_{cl}$ is the balanced hyper-parameter for CL loss.


\section{Experiments}

\subsection{Datasets}

\textbf{NTU RGB+D}.
NTU RGB+D \cite{ntu-2016} is a widely used dataset containing $56,880$ samples. $40$ participants are invited to perform $60$ actions including daily behaviors and health-related actions. Each action is performed by $1$ or $2$ people. The human skeleton is presented by $25$ 3D joints, which are captured by $3$ Microsoft Kinect v2 cameras with different horizontal angle settings. It provides two benchmarks: (1) Cross-Subject (X-Sub): the dataset is divided according to the subjects. The training set consists of 20 subjects while the testing set consists of other 20 subjects. (2) Cross-View (X-View): the dataset is split by the camera views. They select camera views 2 and 3 to construct the training data while camera view 1 is used for testing.

\textbf{NTU RGB+D 120}.
NTU RGB+D 120 \cite{ntu120-2019} extends NTU RGB+D with extra $57,367$ samples by introducing new $60$ action classes, making it the largest skeleton based action recognition dataset.  In total, it collects $113,945$ skeleton sequences over $120$ different classes performed by $106$ participants. It also increases the number of camera setups to $32$ by using different places and backgrounds. Two evaluation protocols are recommended: (1) Cross-Subject (X-Sub): samples from $56$ subjects are selected to form the training set, and the reaming $50$ subjects are used for testing. (2) Cross-Set (X-Set): samples with even setup IDs are used for training, while samples with odd setup IDs are used for testing.

\textbf{NW-UCLA}.
Northwestern-UCLA dataset \cite{N-UCLA-2014} contains $1494$ video clips performed by 10 volunteers. $3$ Kinect cameras are used to capture 3D skeletons with $20$ joints from multiple views. Totally $10$ action categories are covered. We adopt the evaluation protocols recommended by the author: training data comes from the first two cameras, while testing data is from the other camera. 

\begin{table}[t]
\centering
  \caption{Ablation studies of our method on NTU-RGB+D 120 dataset under the X-Sub setting with the joint input modality.}
  \label{tab::Ablation}
  \begin{tabular}{l c c}
  \toprule
  \textbf{Method} & \textbf{Params.} & \textbf{Acc (\%)} \\
  \midrule
  \midrule
  Baseline & 1.46M & 84.5 \\
  \midrule
  + CL Loss & 1.53M & $85.0^{\uparrow 0.5}$ \\
  + ST Decouple & 1.59M & $85.3^{\uparrow 0.8}$ \\
  + ML Refine & 1.61M & $84.7^{\uparrow 0.2}$ \\
  \midrule 
  Ours & 1.99M & $85.5^{\uparrow 1.0}$ \\
    
  \bottomrule
\end{tabular}
\end{table}

\begin{table}[t]
\centering
  \caption{Hyper-parameter exploration of our proposed method on NTU-RGB+D 120 dataset under the X-Sub setting with the joint input modality. The best one is in \textbf{bold}.}
  \label{tab::Params}
  \begin{tabular}{l l l l l c}
  \toprule
  \bm{$w_{cl}$} & \bm{$\lambda_1$} & \bm{$\lambda_2$} & \bm{$\lambda_3$} & \bm{$\lambda_4$} & \textbf{Acc (\%)} \\
  \midrule
  \midrule
  1 & 0 & 0 & 0 & 1 &  84.4 \\
  0.1 & 0 & 0 & 0 & 1 & 85.3 \\
  0.01 & 0 & 0 & 0 & 1 &  85.0 \\
  \midrule
  \multirow{8}{*}{\textbf{0.1}} & 1 & 1 & 1 & 1 & 84.5 \\
  & 1 & 0.2 & 0.2 & 1 & 84.7 \\
  & 1 & 0.5 & 0.2 & 0.1 & 84.1 \\
  & 0.1 & 0.1 & 1 & 1 & 85.2 \\
  & 0.1 & 0.1 & 0.1 & 1 & 85.1 \\
  & 0.1 & 0.2 & 0.2 & 1 & 85.4 \\
  & \textbf{0.1} & \textbf{0.2} & \textbf{0.5} & \textbf{1} & \textbf{85.5} \\
  \bottomrule
\end{tabular}
\end{table}

\subsection{Implementation Details}

We adopt \cite{CTR-GCN-2021} as the backbone and implement the proposed method with the PyTorch deep learning framework. All experiments are conducted on one RTX 2080Ti GPU. The Stochastic Gradient Descent (SGD) optimizer is employed with a momentum of $0.9$ and a weight decay of $0.0004$ to train the models. In the first 5 epochs, we apply a warmup strategy for stable training. The initial learning rate is set to $0.1$ and we decrease it at epoch 35 and 55 with a factor of $0.1$. We train all models with 70 epochs and select the best performance. The base channel $C$ is set to $64$ and the hidden channel $C_h$ is set to $256$. The hyper-parameters in our methods are set as: $\lambda_1 = 0.1, \lambda_2 = 0.2, \lambda_3 = 0.5, \lambda_4 = 1, w_{cl} = 0.1$. For NTU RGB+D and NTU RGB+D 120, we follow the data preprocessing in \cite{SGN2020} and set the batch size to 64. All samples are resized to $64$ frames. For NW-UCLA, we follow the data preprocessing in \cite{Shift-GCN2020} and set the batch size to 16.

\subsection{Ablation Study}
We conduct ablation studies and evaluate the different hyper-parameter settings on the X-Sub benchmark of NTU RGB+D 120 dataset to verify the effectiveness of the proposed module.

The results of ablation studies are displayed in Table~\ref{tab::Ablation}. We remove all additional heads and train the network with CE loss to build the baseline. We also divide the proposed module into different sub-modules and design $3$ variants: (1) CL Loss: we directly employ the CL loss to refine features from the last layer without any additional operations. (2) ST Decouple: we decouple the features into spatial and temporal components before refinement. (3) ML Refine: we impose the refinement on proposed multi-level stages in the training pipeline. It can be seen that all these sub-modules can improve the performance of the baseline. Among them, the contributions from CL loss and ST Decouple are relatively dominant. Moreover, when combining all of them, the result becomes better. We also report the count of trainable parameters of different models. The extra cost of parameters may increase the time of the training procedure but does not affect the inference stage.

We analyze the configurations on the hyper-parameters of our method, and the results are available in Table~\ref{tab::Params}. First, we try $3$ different values of $w_{cl}$ to find the balance between the CL loss and CE loss with the fixed combination of $\lambda_1 = \lambda_2 = \lambda_3 = 0, \lambda_4 = 1$ for an efficient experiment. It seems that bigger $w_{cl}$ may hurt the performance while too small values only provide a little improvement. Then, we try more combinations of $\lambda_i$ to balance the importance of different stages. From the results, we can observe that giving higher weight to the previous layers may obtain negative influence and increase the importance gradually from the early stage to the last stage and thus lead to an optimal result. It is concluded that the refinement of the high-level features from the final stage plays a major role and the low-level features provide the auxiliary effects. Finally, we choose the configuration of $\lambda_1 = 0.1, \lambda_2 = 0.2, \lambda_3 = 0.5, \lambda_4 = 1, w_{cl} = 0.1$ for the following experiments.

\begin{table}[t]
\centering
  \caption{Performance of our proposed method using different GCN-based backbones on NTU-RGB+D 120 dataset with the joint input modality.}
  \label{tab::Backbones}
  \begin{tabular}{l c c c}
  \toprule
  \multirow{2}{*}{ \textbf{Method} } & \multirow{2}{*}{ \textbf{Params.} } & \multicolumn{2}{c}{ \textbf{NTU-RGB+D 120} } \\
     		& & \textbf{X-Sub (\%)} & \textbf{X-Set (\%)} \\
  \midrule
  \midrule
  ST-GCN \cite{ST-GCN-2018} & 2.11M & 83.4 & 85.1 \\
  ~~+ FR Head & 2.65M & $84.4^{\uparrow 1.0}$ & $86.5^{\uparrow 1.4}$ \\
  \midrule
  2s-AGCN \cite{2s-AGCN2019} & 3.80M & 84.3 & 85.9 \\
  ~~+ FR Head & 4.33M & $84.6^{\uparrow 0.3}$ & $86.6^{\uparrow 0.7}$ \\
  \midrule
  CTR-GCN \cite{CTR-GCN-2021} & 1.46M & 84.5 & 86.6 \\
  ~~+ FR Head & 1.99M & $85.5^{\uparrow 1.0}$ & $87.3^{\uparrow 0.7}$ \\
  \midrule
  TCA-GCN \cite{TCA-GCN-2022} & 5.65M & 85.0 & 86.3 \\
  ~~+ FR Head & 6.18M & $85.2^{\uparrow 0.2}$ & $87.4^{\uparrow 1.1}$ \\
  \midrule
  HD-GCN \cite{HDGCN-2022} & 1.68M & 85.1 & 87.2 \\
  ~~+ FR Head & 2.21M & $85.4^{\uparrow 0.3}$ & $87.7^{\uparrow 0.5}$ \\
  \bottomrule
\end{tabular}
\end{table}

\begin{table}[t]
\centering
	\caption{Accuracy (\%) on different difficult level actions for NTU-RGB+D 120 dataset under the X-Sub setting with the joint input modality.}
	\label{tab:Hard}
	\begin{tabular}{l c c c}
		\toprule
		\multirow{2}{*}{Method} & \multicolumn{3}{c}{NTU-RGB+D 120} \\
		& Hard & Medium & Easy \\
		\midrule
		\midrule
		ST-GCN \cite{ST-GCN-2018} & 57.4 & 80.9 & 94.7 \\
		2s-AGCN \cite{2s-AGCN2019} & 58.9 & 82.0 & 95.0 \\
		CTR-GCN \cite{CTR-GCN-2021} & \underline{59.6} & \underline{82.4} & \underline{95.1} \\
		Ours & \textbf{61.6} & \textbf{83.3} & \textbf{95.7} \\
		\midrule
		$\Delta$ & 2.0 & 0.9 & 0.6 \\ 
		\bottomrule
	\end{tabular}	
\end{table}

\subsection{Combined with Other Backbones}
Our proposed module is plug-and-play and compatible with most GCN-based backbones. To examine its universality, we apply it to $5$ widely used GCN-based backbones \cite{ST-GCN-2018, 2s-AGCN2019, CTR-GCN-2021, TCA-GCN-2022, HDGCN-2022} and evaluate them on the X-Sub and X-Set of NTU RGB+D 120 dataset. For fair comparisons, we reimplement them and use the same data preprocessing without the multi-stream fusion. Table~\ref{tab::Backbones} reports the performance and number of parameters of different methods. It is observed that all models obtain an obvious gain of accuracy by employing the FR Head. The improvement is around $1.0\%$. In most cases, the models with lower accuracy are improved more than those with higher initial accuracy. The reason behind it may be that our modules utilize the knowledge from the misclassified samples. The lower accuracy of means the misclassified samples are more sufficient. The additional count of parameters introduced by the FR Head is around 0.5M and can be ignored in the inference stage.

\subsection{Performance on Ambiguous Actions}
We spilt NTU-RGB+D 120 dataset into 3 subsets with different difficulty levels. Specifically, according to the results of CTR-GCN \cite{CTR-GCN-2021}, we gather actions whose accuracy is lower than 70\% as Hard Level, between 70\% and 90\% as Medium Level, and over 90\% as Easy Level. The results are displayed in Table~\ref{tab:Hard}. The experiment is under the X-Subsetting with only the joint input modality. Because ambiguous actions are quite similar and easy to be misclassified, these actions usually fall into Hard Level. From the results, we can see that our method makes a great improvement in Hard Level actions, which demonstrates the ability to distinguish those ambiguous actions.

\begin{figure}[t]
  \centering
   \includegraphics[width=\linewidth]{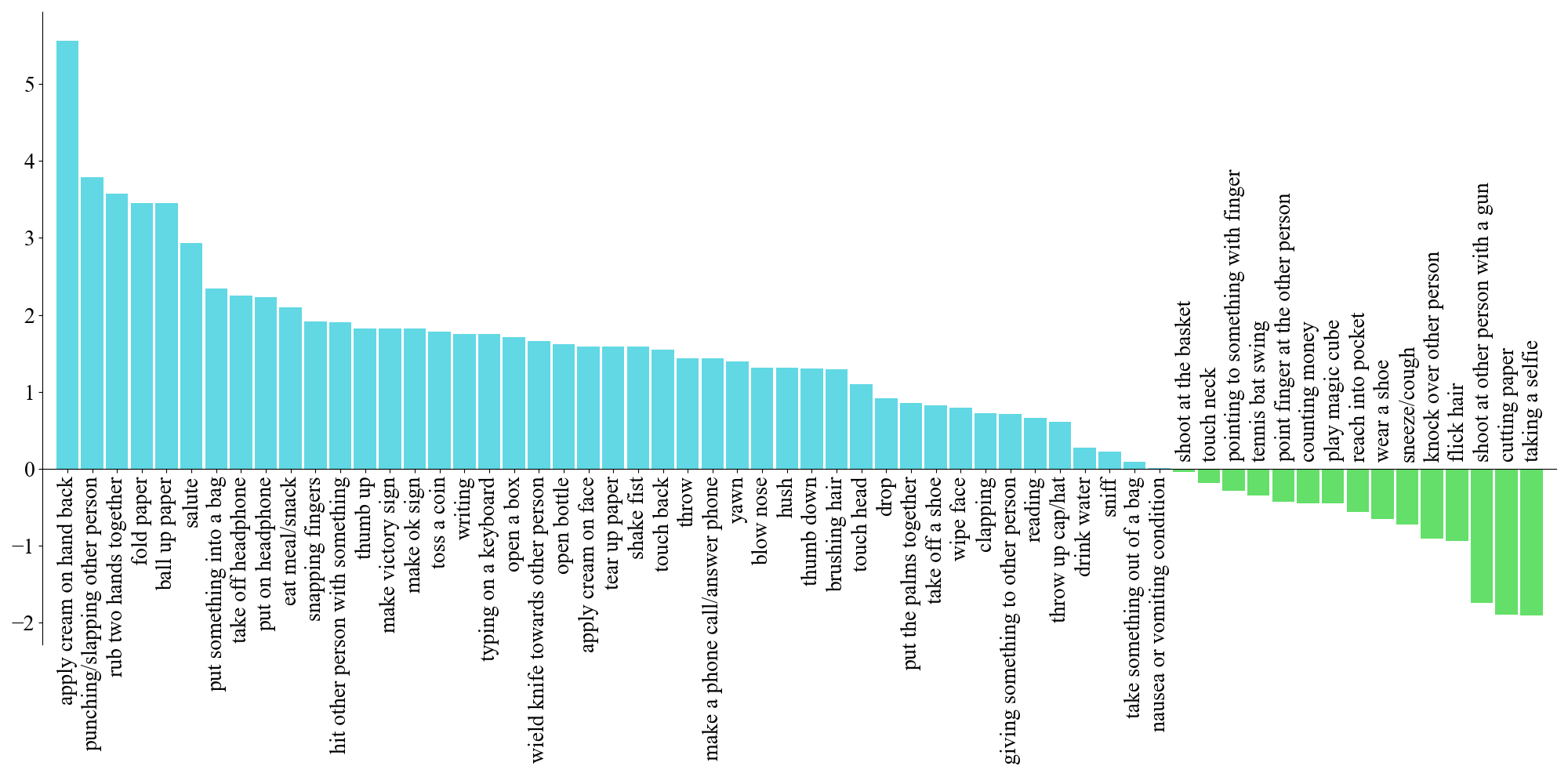}
   \caption{The group-wise accuracy difference (\%) between our method and CTR-GCN \cite{CTR-GCN-2021} on ambiguous actions for NTU-RGB+D 120 dataset under the X-Sub setting with the joint input modality.}
   \label{fig:ambiguous}
\end{figure}

Furthermore, we define ambiguous groups, which collect several related ambiguous actions to verify the performance of ambiguous samples. We first pick a class as an anchor class, for example, "writing". Then we gather the misclassified samples on "writing" and obtain the top-3 actions with the highest frequency, like "reading", "typing on a keyboard" and "playing with phone". These 4 actions will be constructed as an ambiguous group. The group-wise accuracy will be the average accuracy of all actions included by the group. Here, we randomly pick 60 anchor actions and constructed corresponding ambiguous groups from NTU-RGB+D 120 dataset. We compare our results with a SOTA model CTR-GCN \cite{CTR-GCN-2021} and display the results in Fig.~\ref{fig:ambiguous}. Our method gains great improvement in most ambiguous groups. 

\begin{figure}[t]
  \centering
   \includegraphics[width=0.24\linewidth]{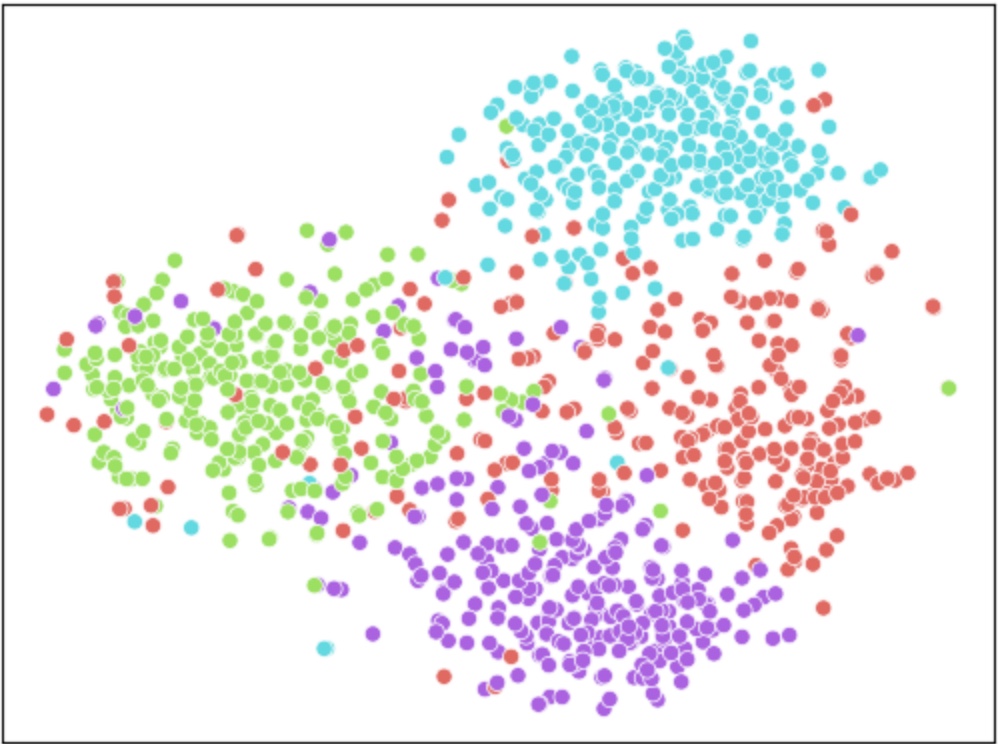}
   \includegraphics[width=0.24\linewidth]{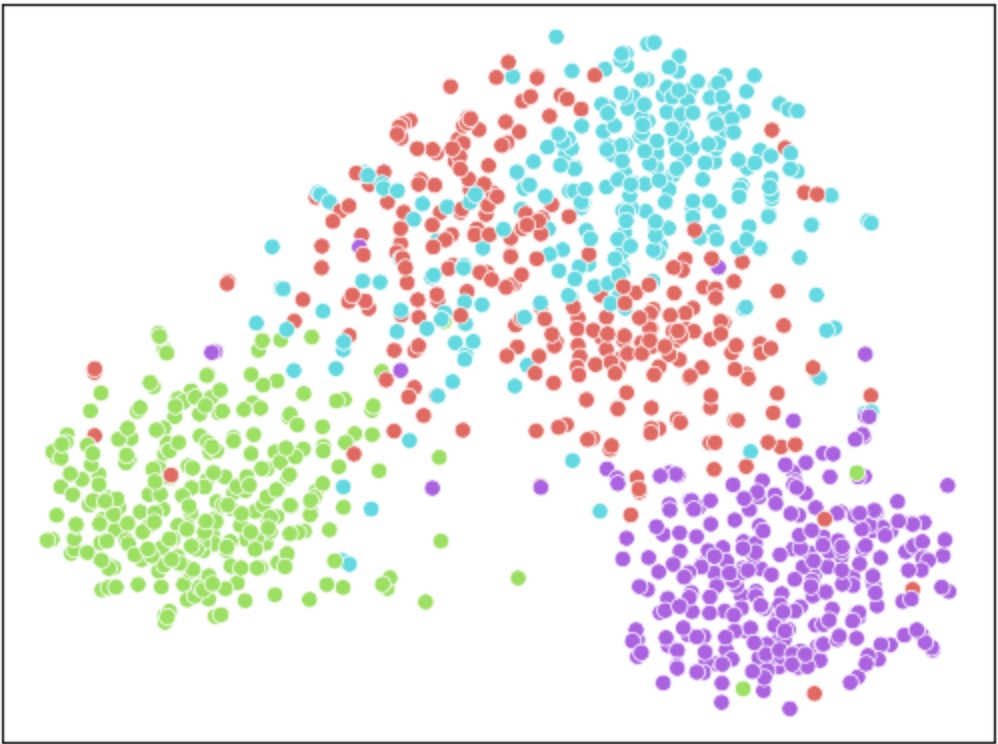}
   \includegraphics[width=0.24\linewidth]{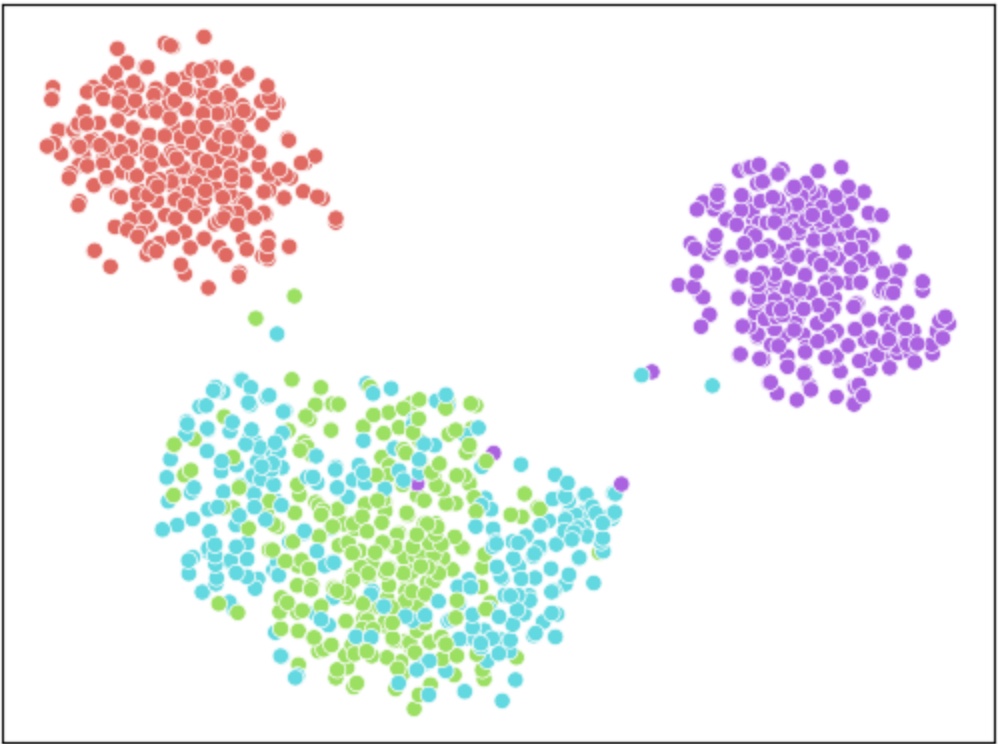}
   \includegraphics[width=0.24\linewidth]{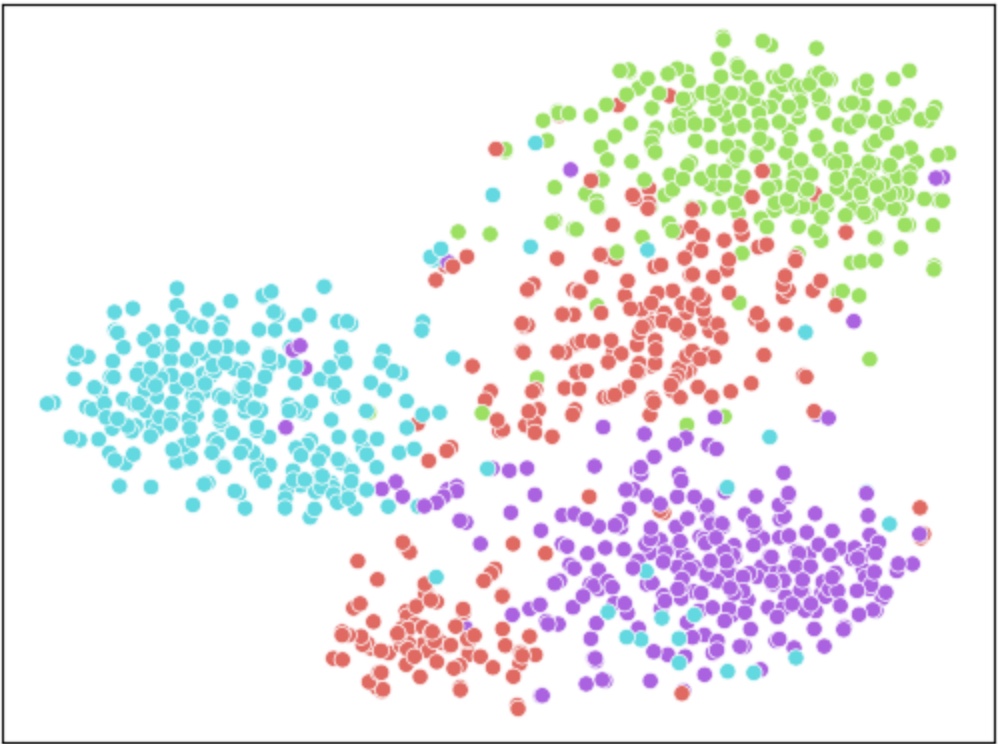}
	   
	\vspace{2pt}	   
	   
   \includegraphics[width=0.24\linewidth]{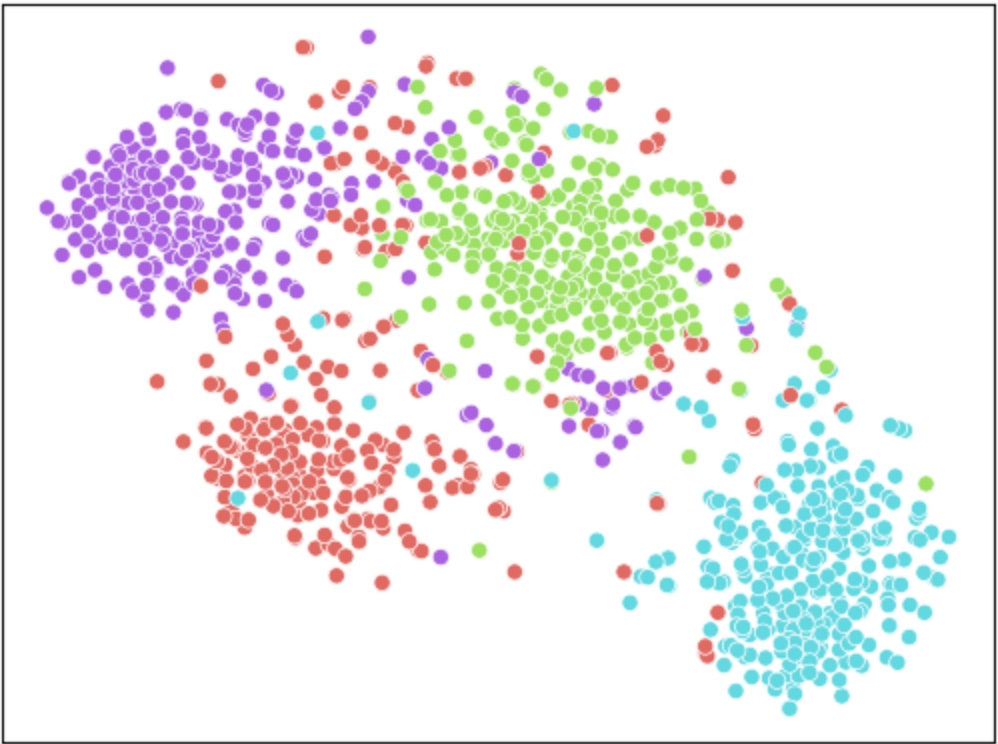}
   \includegraphics[width=0.24\linewidth]{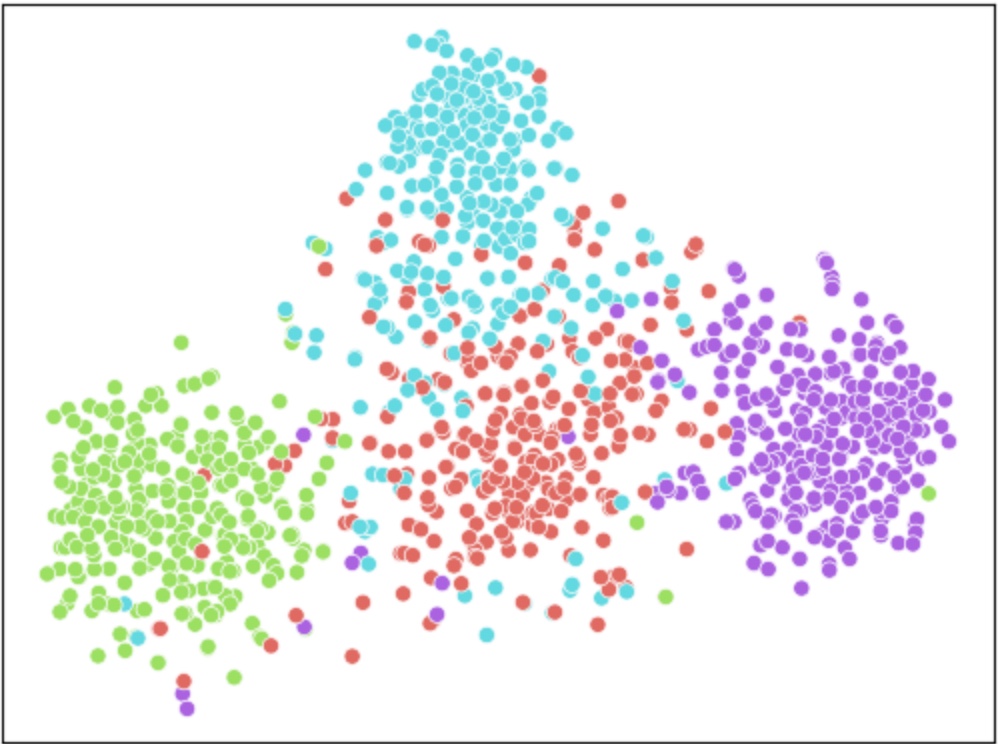}
   \includegraphics[width=0.24\linewidth]{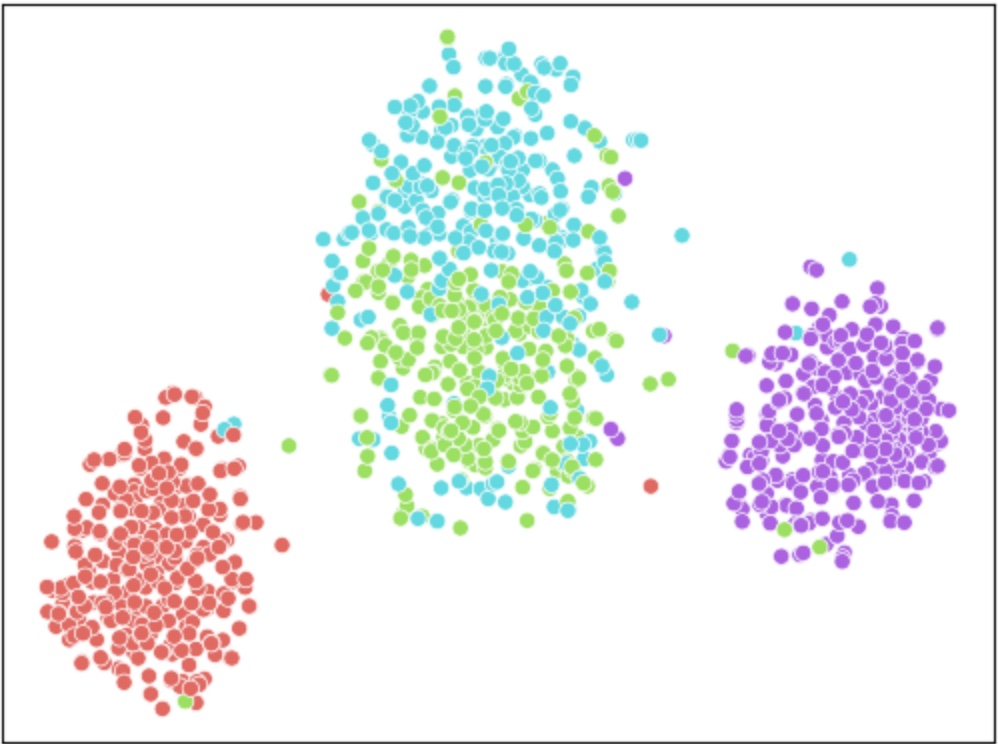}
   \includegraphics[width=0.24\linewidth]{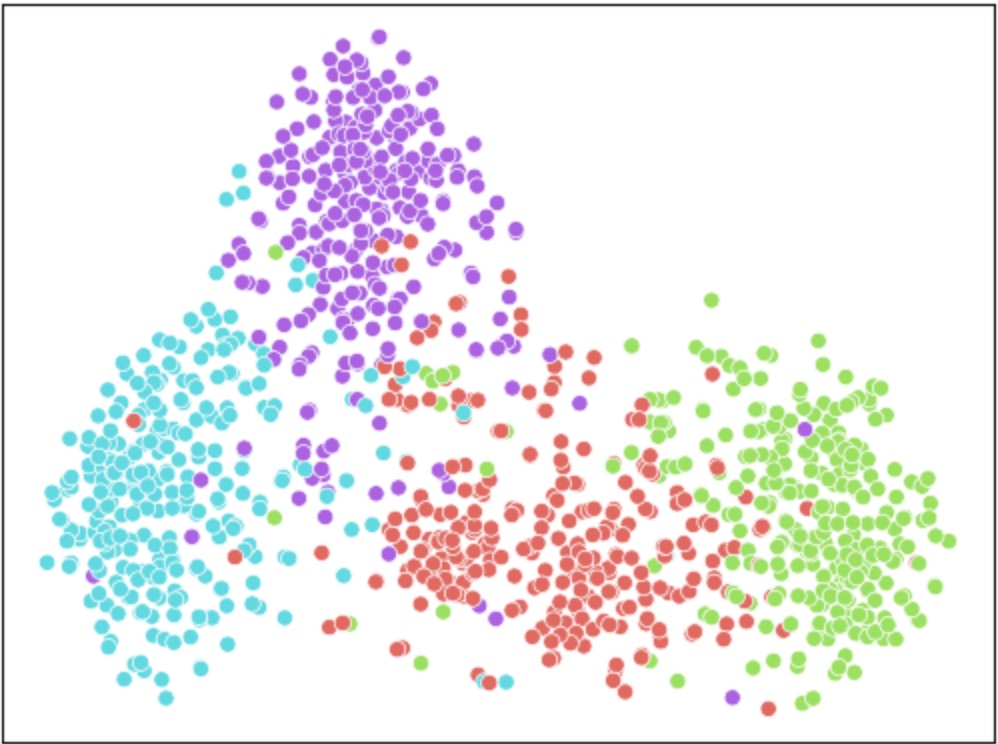}
   
%
%
   
   \caption{Visualization of latent representation by t-SNE for ambiguous groups from NTU RGB+D 120 dataset. Different colors indicate different classes. The upper one is from the CTR-GCN \cite{CTR-GCN-2021}, while the bottem one is from our method.}
   \label{fig:tsne}
\end{figure}

\begin{table*}[t]
\centering
  \caption{Performance comparison of skeleton-based action recognition in top-1 accuracy (\%). The best one is in \textbf{bold} and the second one is \underline{underlined}.}
  \label{tab::ALL}
  \begin{tabular}{r r c c c c c}
    \toprule
     \multirow{2}{*}{ \textbf{Method} } & \multirow{2}{*}{ \textbf{Publication} } & \multicolumn{2}{c}{ \textbf{NTU RGB+D} } & \multicolumn{2}{c}{ \textbf{NTU RGB+D 120} } & \multirow{2}{*}{\textbf{NW-UCLA}} \\
     		& & \textbf{X-Sub} & \textbf{X-View} & \textbf{X-Sub} & \textbf{X-Set} \\
    \midrule
    ST-GCN \cite{ST-GCN-2018} & AAAI2018 & 81.5 & 88.3 & - & - & - \\
    Ind-RNN \cite{Ind-RNN2018} & CVPR2018 & 81.8 & 88.0 & - & - & - \\
    RotClips+MTCNN \cite{RotCLips+MTCNN2018} & TIP2018 & - & - & 62.2 & 61.8 & - \\
    2s-AGCN \cite{2s-AGCN2019} & CVPR2019 & 88.5 & 95.1 & 82.9 & 84.9 & - \\
    AGC-LSTM \cite{AGC-LSTM2019} & CVPR2019 & 89.2 & 95.0 & - & - & 93.3 \\
    DGNN \cite{DGNN2019} & CVPR2019 & 89.9 & 96.1 & - & - & - \\ 
    PA-ResGCN-B19 \cite{PA-ResGCN2020} & ACMMM2020 & 90.9 & 96.0 & 87.3 & 88.3 & - \\
    Dynamic GCN \cite{DynamicGCN2020} & ACMMM2020 & 91.5 & 96.0 & 87.3 & 88.6 & - \\
    SGN \cite{SGN2020} & CVPR2020 & 89.0 & 94.5 & 79.2 & 81.5 & - \\
    Shift-GCN \cite{Shift-GCN2020} & CVPR2020 & 90.7 & 96.5 & 85.9 & 87.6 & 94.6 \\
    MS-G3D \cite{MS-G3D-2020} & CVPR2020 & 91.5 & 96.2 & 86.9 & 88.4 & - \\
    DDGCN \cite{DDGCN2020} & ECCV2020 & 91.1 & \textbf{97.1} & - & - & - \\
    DC-GCN+ADG \cite{DC-GCN+ADG2020} & ECCV2020 & 90.8 & 96.6 & 86.5 & 88.1 & 95.3 \\
    MST-GCN \cite{MST-GCN-2021} & AAAI2021 & 91.5 & 96.6 & 87.5 & 88.8 & - \\
    Skeletal-GNN \cite{Skeletal-GNN-2021} & ICCV2021 & 91.6 & 96.7 & 87.5 & 89.2 & - \\
    CTR-GCN \cite{CTR-GCN-2021} & ICCV2021 & 92.4 & 96.8 & \underline{88.9} & \underline{90.6} & \underline{96.5} \\
    STF \cite{STF2022} & AAAI2022 & \underline{92.5} & \underline{96.9} & \underline{88.9} & 89.9 & - \\
    Ta-CNN \cite{Ta-CNN2022} & AAAI2022 & 90.4 & 94.8 & 85.4 & 86.8 & 96.1 \\
    EfficientGCN-B4 \cite{EfficientGCN2022} & TPAMI2022 & 91.7 & 95.7 & 88.3 & 89.1 & - \\
    \midrule
    \textbf{Ours} & - & \textbf{92.8} & 96.8 & \textbf{89.5} &  \textbf{90.9} & \textbf{96.8} \\
    \bottomrule
\end{tabular}
\end{table*}

We randomly pick some ambiguous groups and visualize the distribution of them in the feature space using t-SNE. As mentioned before, each ambiguous group contains four classes, including an anchor class and three ambiguous classes. We compare our method with CTR-GCN \cite{CTR-GCN-2021}. From Fig.~\ref{fig:tsne} we can see that our model obtains a more discriminative representation resulting in a compact clustering.

\subsection{Comparison with the State-of-the-Art}
In this section, we conduct a comparison with the state-of-the-art methods on NTU RGB+D 120, NTU RGB+D, and NW-UCLA datasets to demonstrate the competitive ability of our proposed module. The quantitative results are displayed in Table~\ref{tab::ALL}. It is noted that most of the state-of-the-art methods employ a multi-stream fusion framework. For a fair comparison, we follow the same framework as \cite{CTR-GCN-2021, TCA-GCN-2022}. We make a fusion with the results from four modalities including joint, bone, joint motion, and bone motion as the final report result.

It is observed that our methods outperform most existing methods on these three datasets. On both settings of NTU-RGB+D 120, X-Sub of NTU-RGB+D, and NW-UCLA datasets, our model obtains the best results. On X-View of NTU-RGB+D, our model reaches state-of-the-art results with a reasonable gap between the best one, which demonstrates the great potential of our proposed module. Notably, our method is the first to propose a way to solve ambiguous actions, which are very important in skeleton based action recognition. 

\section{Conclusion}
In this paper, we present a novel feature refinement module equipped with contrastive learning to solve the ambiguous actions for skeleton based action recognition. Multi-level features extracted from GCN-based backbone are leveraged and enhanced on both the spatial and temporal dimensions. The contrastive learning is conducted with the samples with high confidence and calibrated by the FP and FN samples to make full use of the misclassified actions. 

The extensive experiments demonstrate the effectiveness of the proposed module to distinguish the confusing categories and the university to be compatible with most GCN-based backbones. On three widely used benchmarks, our proposed method obtains satisfactory results and outperforms those state-of-the-art methods. 

\textbf{Discussion}.
Despite the performance of our proposed module on three public large-scale datasets, the ambiguous actions in a few-shot setting with insufficient data remain to be explored. We will concentrate on it in our future work. In addition, there are some potential negative societal impacts to be considered. Our method may be applied in some controversial fields, such as surveillance. Besides, applying our module will introduce extra training costs, which should be discussed in the carbon emission problem.



{\small
\bibliographystyle{ieee_fullname}
\bibliography{egbib}
}

%

\end{document}